\documentclass[letterpaper, 10 pt, conference]{ieeeconf}  

\IEEEoverridecommandlockouts                              

\usepackage[acronym]{glossaries}
\usepackage{graphicx}
\usepackage[bookmarks=true]{hyperref}
\usepackage{siunitx}
\usepackage{amsmath}
\newcommand{\mathbbm}[1]{\text{\usefont{U}{dsrom}{m}{n}#1}}
\usepackage{mathtools}
\mathtoolsset{showonlyrefs=true}
\usepackage{multicol}
\makeglossaries{}
\newacronym{grone}{GRoNe}{Grande Region rObotique aerienNE}
\newacronym{arma}{ARMA}{Auto Regressive Moving Average}
\newacronym{ros}{ROS}{Robotic Operating System}
\newacronym{mpc}{MPC}{Model Predictive Control}
\newacronym{mse}{MSE}{Mean Squared Error}
\newacronym{mlp}{MLP}{Multi Layer Perceptron}
\newacronym{lstm}{LSTM}{Long Short-Term Memory}
\newacronym{mppi}{MPPI}{Model Predictive Path Integral}
\newacronym{icp}{ICP}{Iterative Closest Point Algorithm}
\newacronym{siso}{SISO}{single-input single-output}
\newacronym{mimo}{MIMO}{multiple-input multiple-output}
\newacronym{rtk}{RTK}{Real Time Kinematic}
\newacronym{rl}{RL}{Reinforcement Learning}
\newacronym{ppk}{PPK}{Post Processing Kinematic}
\newacronym{nn}{NN}{Neural Network}
\newacronym{per}{PER}{Prioritized Experience Replay}
\newacronym{grad}{GRAD}{upper-bound gradient prioritization}
\newacronym{usv}{USV}{Unmanned Surface Vehicle}

\title{\LARGE \bf
How To Train Your HERON
}

\author{Antoine Richard$^{1,2}$, St\'ephanie Aravecchia$^{2}$, Thomas Schillaci$^{1}$, Matthieu Geist$^{3}$, C\'edric Pradalier$^{2}$ 
\thanks{$^{1}$ are with Georgia Institute of Technology, USA
        {\tt\small first.last@gatech.edu}}%
\thanks{$^{2}$ are with UMI2958 GT-CNRS, 2 rue Marconi, 57070 Metz, France}%
\thanks{$^{3}$ is with Google Research, Brain Team}%
}

\begin{document}

\maketitle
\thispagestyle{empty}
\pagestyle{empty}

\begin{abstract}
In this paper we apply Deep Reinforcement Learning (Deep RL) and Domain Randomization to solve a navigation task in a natural environment relying solely on a 2D laser scanner. We train a  model-based RL agent in simulation to follow lake and river shores and apply it on a real Unmanned Surface Vehicle in a zero-shot setup. We demonstrate that even though the agent has not been trained in the real world, it can fulfill its task successfully and adapt to changes in the robot's environment and dynamics. Finally, we show that the RL agent is more robust, faster, and more accurate than a state-aware Model-Predictive-Controller. Code, simulation environments, pre-trained models, and datasets are available at ~\url{https://github.com/AntoineRichard/Heron-RL-ICRA.git}.
\end{abstract}

\section{Introduction}
Autonomous navigation in natural environments is critical in areas such as agriculture, inspection or environment monitoring. In these tasks, robots have to perform actions in complex and unstructured scenes,  constantly changing. This requires them to have an in-depth understanding of their surroundings and own dynamics. 
%
To solve such tasks, a robot needs either 
a behavior model or a local planner. When using a local planner, the controller needs a dynamic model to follow the path and optionally an interaction model to know how changes in the environment impact the system.

Nowadays, most of the controllers use this kind of design~\cite{lenain2006high}. However, as they rely on an accurate depiction of the robot dynamics, they require to measure the state of the system precisely. This means that the robot is carrying an expensive suite of sensors to acquire its state. Even though these approaches perform well, their cost, and the size of the sensors, make them hard to apply on small embedded systems. Furthermore, some sensors such as RTK-GPSs only work reliably in open-sky areas making them impractical to use in occluded spaces, in forests, or in adverse meteorological conditions. 
Based on these observations, controllers based only on a small number of sensors such as cameras or laser scanners have gained interest over the recent years. The rise of these approaches was facilitated by the fast growth of the reinforcement learning (RL) field~\cite{Arulkumaran_2017}, with its numerous benchmarks and the ever increasing computational resources available on low-power devices. Building on this, an increasing amount of studies lean on RL-based controllers for their robot; among the popular methods, model-free techniques such as A3C~\cite{mnih2016asynchronous} or SAC~\cite{haarnoja2018soft} have been used extensively~\cite{cai2020high, hong2018virtual}. However, these approaches are rarely deployed in the field. 
This could be due to the cumbersomeness of training RL agents, and the issues inherent to field robotics.

In this work we teach an under-actuated Unmanned Surface Vehicle (USV) that exhibits thrust non-linearities to follow lake and river shores using DREAMER~\cite{Hafner2020Dream}, a model-based RL technique. Model-based RL is particularly interesting in robotics as it is more efficient than model-free RL, reducing the amount of interaction with the environment needed to train the model. Another interesting point is that DREAMER, which uses latent imagination, natively learns the dynamics of the system to build its world model. To do so our agent is exclusively provided with measurements from a laser scanner and must navigate at a fixed distance from the shoreline. Please notice this task \emph{is not} a path following task, as we rely on the environment to reactively find a path to follow. To the best of our knowledge this is the first time DREAMER is being applied onto robots.

\begin{figure}
    \centering
    \includegraphics[width=1.0\linewidth]{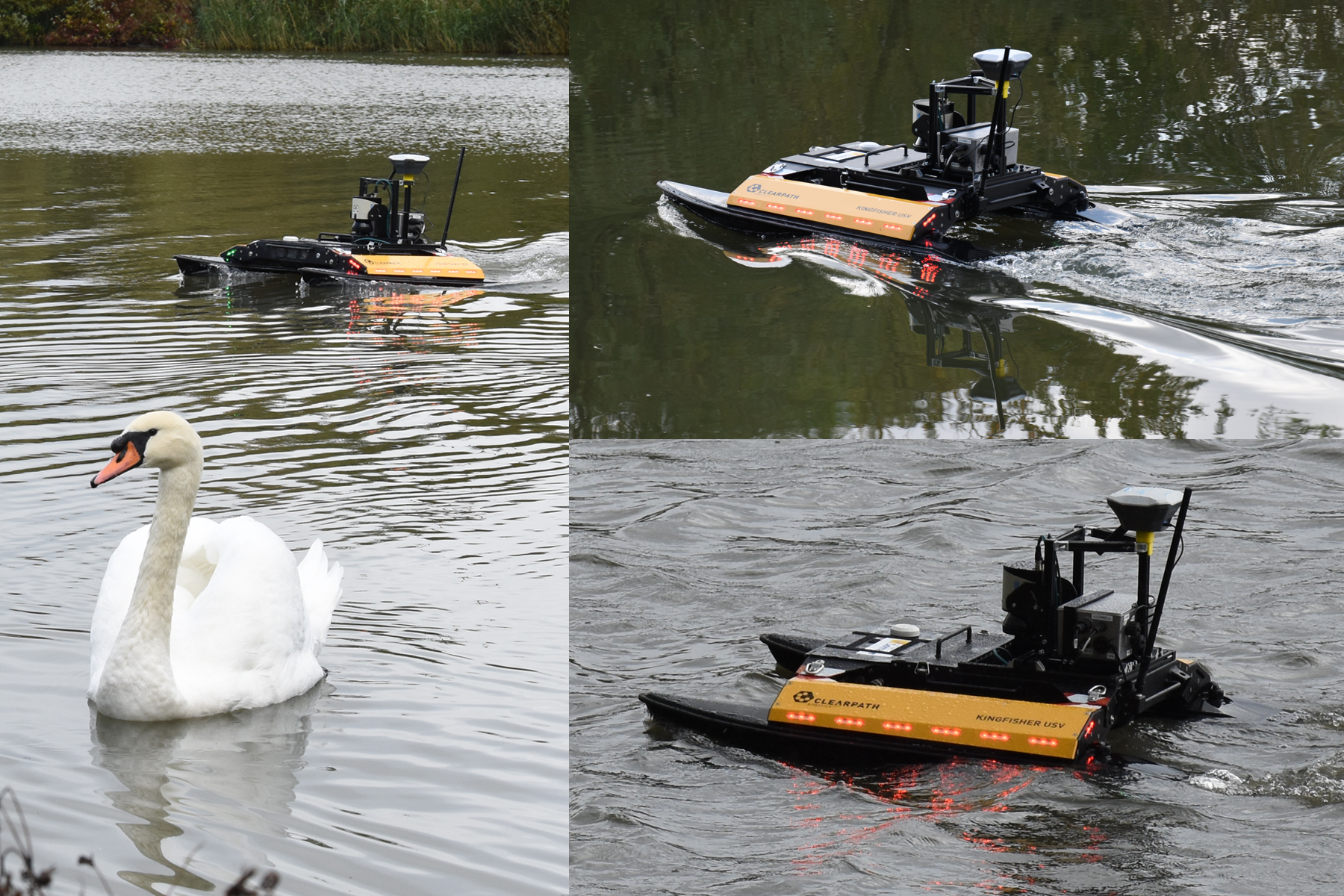} 
    \caption{The Heron dashing around the Symphonie lake.}
    \label{fig:my_label}
\end{figure}

Yet, such RL agents come with disadvantages: they still require a lot of interactions to reach satisfying performances; as they learn by trial and error, they need to be constantly monitored to prevent accidents. A solution to this is either to use a very accurate simulator, or spend time training the model in the real environment which is incredibly time consuming and risky. To alleviate these issues, efforts have been made to create methods that allow to train agents in simulation and deploy them in real-life~\cite{andrychowicz2020learning, hong2018virtual}. Most of them rely either on Domain Randomization (DR)~\cite{DR-state-init} or Domain Adaptation (DA)~\cite{chebotar2019closing}. While DA uses some real world data to adjust its parameters, DR does not. Here, although we apply DR, our study is more focused on what type of data-representation is used by the agent to minimize the use of these methods and deploy the agent without ever training it on the real system.

Our contribution are:
\textbf{1)} training an RL agent solely in simulation to perform a shore-following task and applying it successfully on the real system in different weather conditions including moderate rain, and wind;
\textbf{2)} evaluating different data representations and how they fare in a zero-shot deployment;
\textbf{3)} reporting on lessons learned and giving some advice to practitioners.

\section{Related Work}
Designing algorithms to tackle reactive navigation problems is an active field of research in robotics since the 80s, as detailed in \cite{10.5555/521898}. 
Many of these methods apply learned (RL) policies~\cite{ram1992case}. Actually, RL for policy search in robotics has been studied for decades~\cite{deisenroth2013survey, kober2013reinforcement}, and was applied to solve state-aware problems way before Deep RL came to be. However, those approaches often failed to process directly high dimensional inputs such as images. This is where Deep RL steps in, as neural networks allow them to process complex inputs and act upon them.


\subsection{Deep RL for Perception-Based Control}
The emergence of deep learning has led to breakthroughs in many fields, and RL is no exception. The resulting field, Deep RL, is presented in~\cite{Arulkumaran_2017}. Perception-based control happens to be the very task tackled by one of the most iconic  Deep RL agent, namely DQN~\cite{Mnih2015}, an agent that achieves human-performance on a suite of classic video games. As of today, 
RL agents can be separated into two main families: model-free and model-based agents. 

As their name suggests it, model-free agents do not try to learn  transition or reward models. They are some of the most successful RL algorithms to date~\cite{barth2018distributed, lee2020stochastic, haarnoja2018soft}, including for robotics. For example, \cite{hong2018virtual} uses A3C~\cite{mnih2016asynchronous} to perform a visual servoing task where the agent follows a yellow marker in the image. Yet, one big disadvantage of model-free RL is that it usually requires a huge amount of interactions.



On the other hand, model-based approaches are much more data-efficient but they often lagged behind the best model-free agents. 
As model-based agents learn the transition model and the reward model, they can interact with the environment virtually, mitigating the need for interaction with the real system. However, they then learn an imperfect model, which can deteriorate their performances.
In \cite{hafner2019learning, hafner2019dream}, the model is learned from simulation and used to train the policy. This allows the agents to train the policy for hundreds of billions of time-steps, without interacting with the real system.%




\subsection{Zero Shot Learning \& Domain-Randomization}
Even though data efficiency is getting more and more attention from the RL community, 
training an agent in the real world for millions of steps remains cumbersome. To avoid this, a set of methods grouped under the name Domain-Randomization (DR) have been developed. Robotics has seen an extensive use of DR applied to visual servoing~\cite{DR-viz-ex1, DR-viz-ex2, DR-state-init}. For example, \cite{DR-viz-ex1, DR-viz-ex2} apply DR by making changes in the 3D scenes by randomly changing illumination or textures. This allows training object-detectors and collision detectors that work in the real world without ever being trained in it. 

DR has also been used to change the dynamics of the system for learning robust policies~\cite{DR-state-init, DR-state-ex1, DR-state-ex2}. 
DR does not require real-world data, but training a model under DR takes more time than what it would have if the true system was known.
Finally, \cite{hong2018virtual} trains an A3C agent to follow yellow semantic labels in simulation and shows that the agent transfers seamlessly between the simulation and the real world. This task is akin to a visual servoing task where the robot is steered solely based on the label it has to follow.

In this paper we apply DREAMER~\cite{hafner2019dream}, a model-based RL agent, and use DR to learn a policy that is robust to the perturbation that can occur in the real world.

\section{Method}
\begin{figure}[b]
    \centering
    \includegraphics[width=\linewidth]{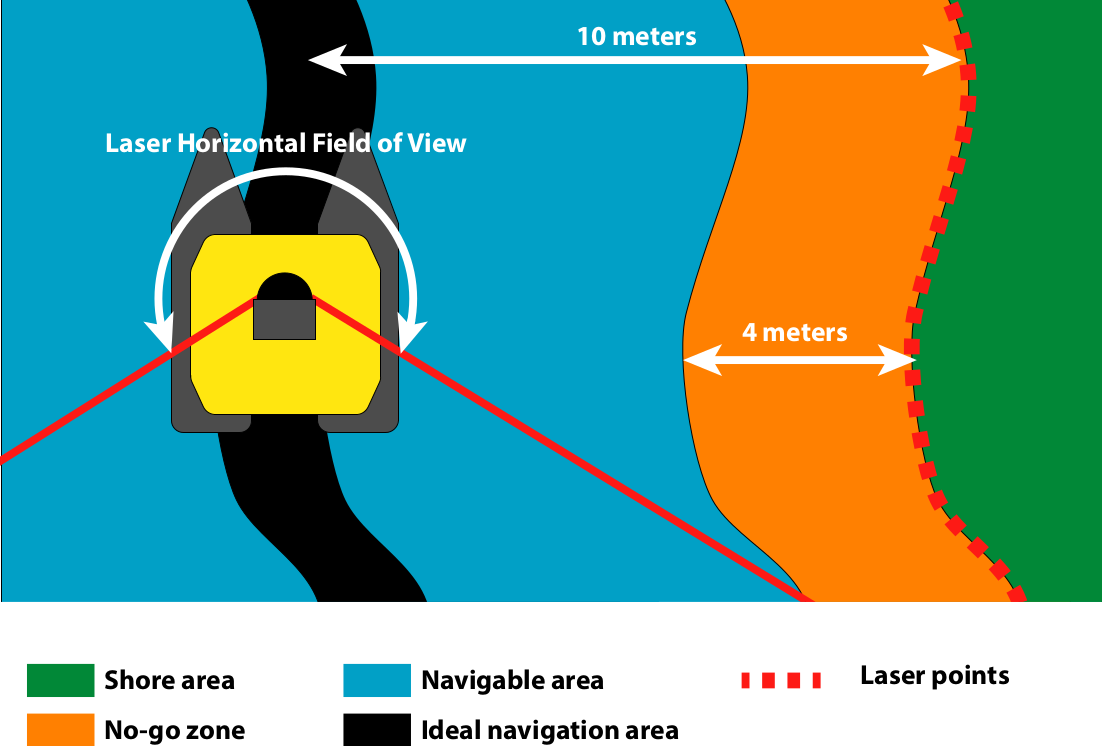}
    \caption{The USV and its shore-following task. (Colors are illustrative.) }
    \label{fig:task}
\end{figure}

This section details how the RL agent is trained, in simulation, to perform a shore-following task, both in simulation and on the lake. To follow the shore, our agent relies only on a SICK laser scanner. Its goals are to remain at a fixed distance from the shore, to maintain a constant velocity, and to turn around the lake in a single direction. Figure \ref{fig:task} shows the robot and the task it has to accomplish.

\subsection{DREAMER}

The RL agent used in this study is articulated around five main quantities. The input from the environment $o$: in our case a laser-scan, or a projection of it. The state $s$ which is a learned latent space that contains the dynamics of the agent and all the information needed for it to construct its representation of the world. The action $a$, a vector in ${\rm I\!R}^2$. The reward $r \in {\rm I\!R}$  that the model tries to maximize and a value $v \in {\rm I\!R}$.
DREAMER itself is built around 3 blocks: a dynamic model parameterized by $\theta$, a policy function by $\phi$ and value function by $\psi$. The dynamic model is the world-model of our agent. It consists of three functions: a representation function: $p_{\theta}(s_t | o_t, a_{t-1}, s_{t-1})$, a transition function $q_{\theta}(s_t | s_{t-1}, a_{t-1})$ and a reward function $q_{\theta}(r_t | s_t)$. This model is trained off-policy on previously recorded interactions of the agent with its environment. To be optimized the world-model requires from the simulation sequences of $(o_t, a_t, r_t)$. The optimization is performed by sequentially predicting a reward $r_t$ from a state $s_t$, reconstructing the observation $o_t$ from a state $s_t$ and minimizing the KL divergence between $s_t$ and $s_{t+1}$ (obtained from $s_{t}$ and $a_{t}$).
The two other blocks of DREAMER, the policy function $q_{\phi}(a_t | s_t)$ and the value functions $v_{\psi}(s_t)$ are learned through a process called latent imagination. This is the process that makes DREAMER highly data-efficient. An accurate depiction of the imagination process can be found in \cite[sec3]{Hafner2020Dream}.
\begin{figure}[h]
    \centering
    \includegraphics[width=0.7\linewidth]{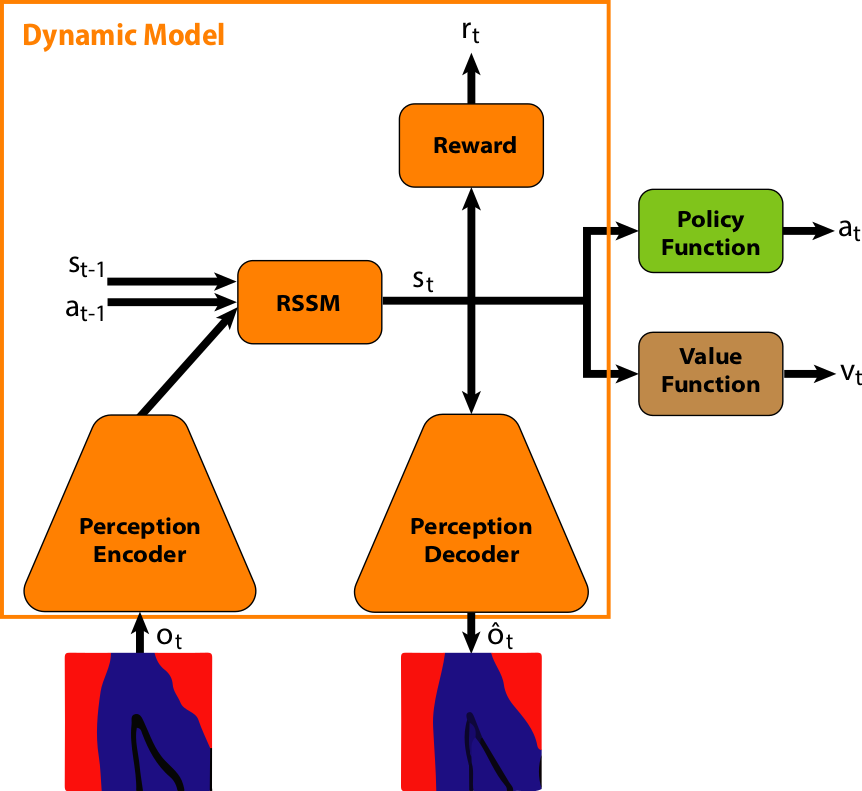}
    \caption{DREAMER's architecture}
    \label{fig:DREAMER}
\end{figure}
In this work we modify the encoder and the decoder of the representation function $p_{\theta}$ of the dynamic model of DREAMER. Initially designed to process images, we adapt them to process 2D laser-scans and study their behavior when transferred from the simulation to the real-world. 
DREAMER's architecture is illustrated in figure~\ref{fig:DREAMER}.

\subsection{Data Representation}
DREAMER's modularity allows us to feed it with an observation and reconstruct something completely different. We study two types of observations and reconstructions.

\subsubsection{Laser Measurements}
The most naive way of using DREAMER with a 2D laser-scan is to change its encoders and decoders to process the laser-scans directly. To do so, we replaced the encoder/decoder based on 2D Convolutional Neural Networks (CNNs) by 1D CNNs. In addition to this modification of DREAMER, we also project the laser-scans to a continuous representation. By default laser-scans are a discontinuous representation of a 2D world: the laser points that are not reflected are set to zero or infinity. To alleviate this issue, we set all the zeros in the laser-scan to an arbitrary large value and compute its inverse. This creates a continuous representation where points close from the USV have a large value and points far from it have a small value. Finally to make the laser more robust to the brutal changes that happen in a natural environment we use an operation that would be similar to a \emph{min\_pool} with a stride of 2. In the end, we remove 7 points on each side such that the final laser-scan has a shape of 1x256x1 making it easy to manipulate using convolution and pooling operations.

\subsubsection{Laser Projection}
While using the laser-scan directly can sound appealing, the transfer between simulation and real environment is difficult.
Indeed, in our natural environment, the movement of the leaves, or small branches, and the partial reflection of laser points due to the semi-transparent character of the vegetation make real and simulated laser-scans very different. 
Thus, we choose to transform the laser-scans into a robust representation. To do so, we create an image representing a local map with a width of 20 meters and a height of 12 meters, the map has a resolution of 1 pixel per 10cm. The origin of the map, the position of the robot, is set 2 meters from the top and 10 meters from the left and its background is blue. Then we convert laser-scans into points in the map and trace 4 meters red circles on top of each point. Finally, we trace a black one meter wide curve 10 meters from the shore to represent the track and resize the map to a 64x64 image. Examples of such images can be seen at the bottom of figure \ref{fig:DREAMER}. This representation is robust to changes and behave similarly in simulation and in the real world, as shown later. 


\subsection{Reward}
%
The reward of our agent is based on two independent metrics: $\Delta_p$, the distance of the agent from the target shore distance $d_{d}$, and $\Delta_v$, the difference between the agent's velocity $v$ and the target linear velocity $v_{d}$. In all our experiments, we set $v_{d} = 1.0$ and $d_{d} = 10$. Furthermore, we penalize the agent if it gets too close to the shore or if it goes backward. 
The reward is $R = 1.25 R_v + 2.5 R_p$, with $R_v = 1 - \|\Delta_v\| \mathbbm{1}_{\{v \geq 0\}} - 0.625 \mathbbm{1}_{\{v<0\}}$ and $R_p = \max(-20, 1 - 2.5\Delta_p)$.
This reward is computed when training the model, and learned as part of it. The agent is trained on the learned reward (estimated solely from the laser-scan measurements, through the learned embedding).


    

\subsection{Environment \& Domain Randomization}

In the real world, the system's dynamics can be impacted by at least two environmental factors: wind and water current. To compensate for those, we add water current to the simulation. More precisely, at the beginning of each episode we draw a water current velocity from a uniform distribution in the range [$0$, $0.4$] $m/s$ and set its direction randomly. 
Additionally, as our real system has gone under significant modifications its characteristics are different from the one used in simulation. To account for that and the approximations of the simulator, we also change the water density at the beginning of every episode. Before the agent starts playing we draw a density from a uniform distribution in the range [$1000$, $2500$] $kg/m^3$. 
Finally, we also make the agent spawn close and far from the shore to make sure it learns how to recover from such events.

\section{Experiments}

\subsection{System}
The simulation and the real experiments were performed using similar systems. The real robot is a Clearpath Robotics Heron~\footnote{For completeness, our robot is a Kingfisher (Heron's previous version).} USV equipped with a SICK LMS111, a 20 meter-range, 270$^{\circ}$ field of view 2D-LiDAR running at 50Hz. The simulated robot is a Clearpath Heron, the new version of the Heron, equipped with the same simulated LiDAR. The weight distribution of the real and simulated systems are different: our real system was adapted to carry an NVIDIA Jetson Xavier and an RTK GPS, the latter being used only for visualization and evaluation purposes.
On the real system, an Intel Atom is used for low-level computations, while the Jetson Xavier at base clocks is used to compute and apply the agent policy. Both computers are running ROS 
with a single master. The main challenge of this system is its inertia: with its current configuration our Heron weight is around 35 kg, and it only has two 400 W motors (one left, one right). 
Hence if an agent wants to take turns correctly it needs to anticipate. 
To run the simulation, infer and train the RL agents we use an NVIDIA RTX 2080Ti and an 8 core CPU server with 16Gb of RAM.
On both the real system and simulation, the RL agent runs using CUDA 10.0, python 2.7, and tensorflow 2.1~\cite{abadi2016tensorflow}. 
The agent model has not been converted using TensorRT or any other DNN compilers and can run easily at 12Hz on the Xavier.

\subsection{Environments}
\begin{figure}[h]
    \setlength{\tabcolsep}{2pt}
    \begin{tabular}{ccc}
        \includegraphics[width=.3\linewidth, height=2.3cm]{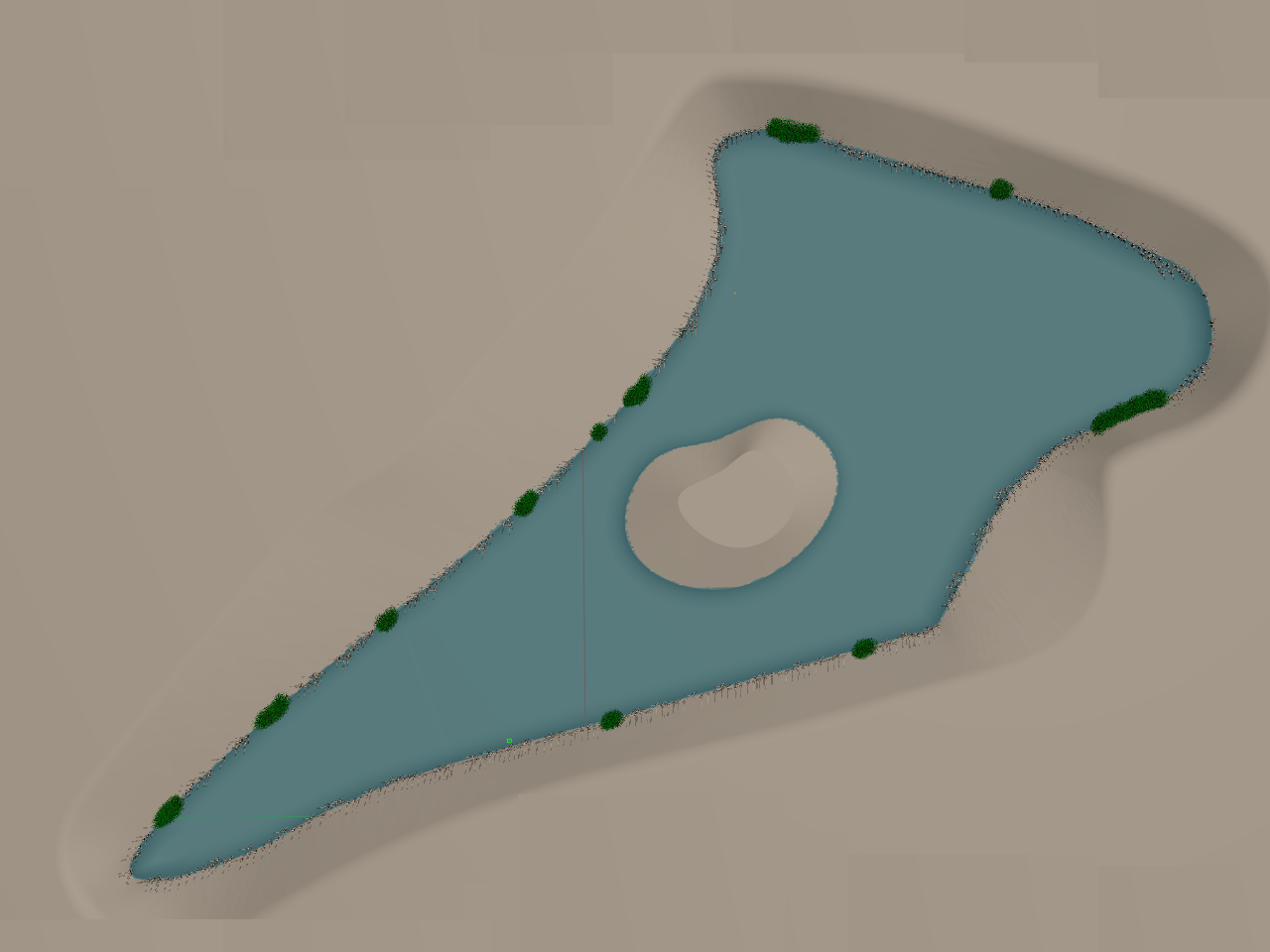}& \includegraphics[width=.3\linewidth, height=2.3cm]{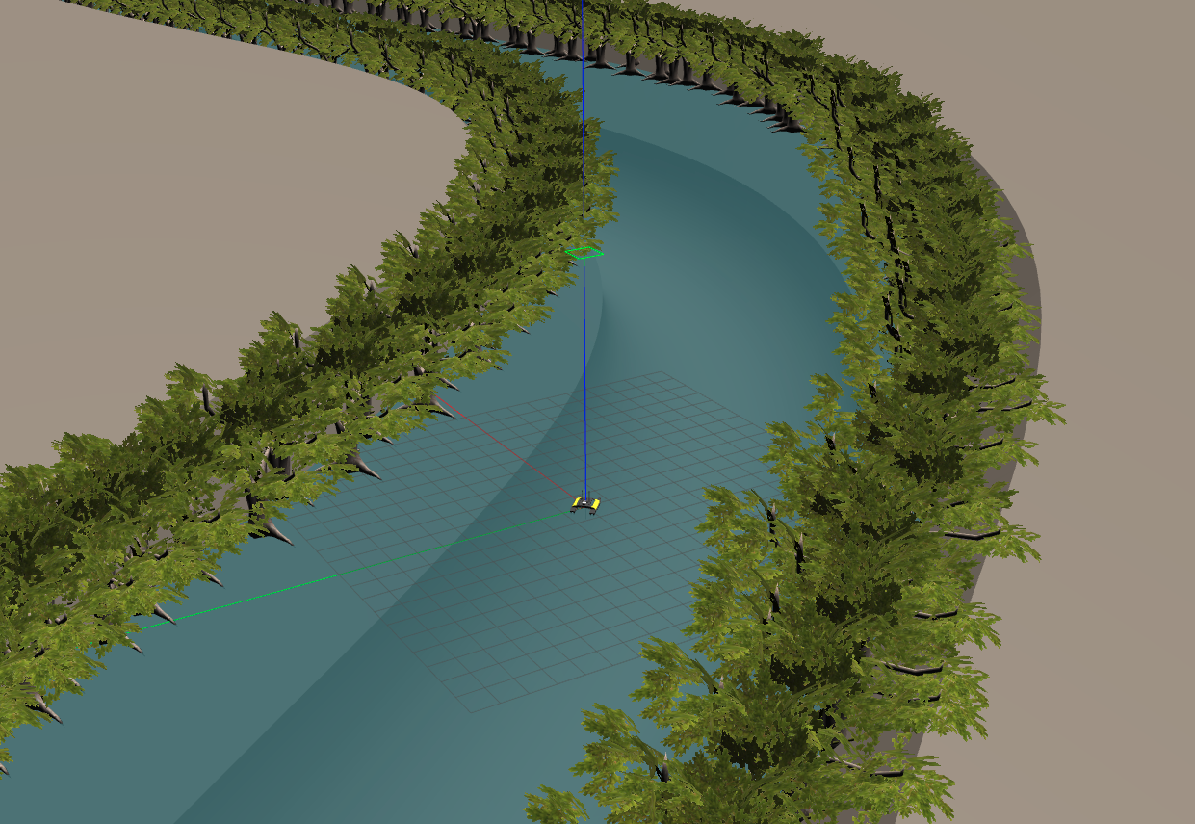}& \includegraphics[width=.3\linewidth, height=2.3cm]{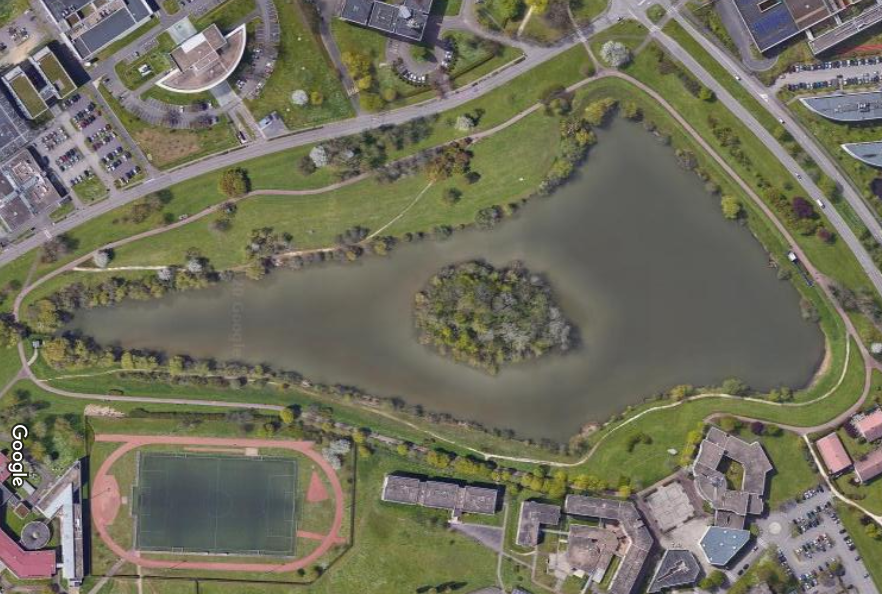}\\
        (a) & (b) & (c)
   \end{tabular}
   \caption{Simulation and Real environments: (a) and (c) are the simulated and real lake (Lac Symphonie, 57000 Metz, France, Google Maps, 2020), (b) is a simulated channel}
   \label{fig:envs}
\end{figure}
\subsubsection{Simulation}
We trained our models in simulation using Gazebo\footnote{\url{gazebosim.org}}, a simulator that supports advanced physics modeling and allows to create custom robots.
The simulation of the USV is done using the ROS packages heron-simulator\footnote{\url{github.com/heron/heron_simulator}} (simulated version of Clearpath's Heron) and uuv-simulator\footnote{\url{github.com/uuvsimulator/uuv_simulator}} (water buoyancy simulation and thrust non-linearities mimicking the real USV behavior). 
To train our agent, we have replicated our real test environment: a lake around which we have planted approximately 1000 trees to create a feature rich and intricate shoreline. We also test our agent in other simulated environments to ensure that it has learned something that can be applied in a wide variety of situations as can be seen in Figure~\ref{fig:envs} (a, b).

\subsubsection{Real Environment}
Our real test environment is a small artificial lake with approximately 1.4 km of shoreline. As can be seen in Figure~\ref{fig:envs} (c) it presents two main complexities, a hairpin bend on the far right of the lake, and a small island that creates a narrow passage. Additionally, we deployed our system in various weather conditions, such as light rain, and wind with average speed of 40 kmph, and gust of wind reaching up to 60 kmph. Furthermore, our lake also features a very active and curious wild-life including swans and ducks that act as moving obstacles.

\subsection{Training parameters}
All our agents are trained for 2000 episodes of 500 steps, the agent plays 12 times per second. This makes for a total of 1.000.000 simulation steps, or about 24 hours of play time. Although we could have played at a higher frequency, we noticed no improvement from 50Hz to 12Hz: we decided to stay at 12Hz. All training parameters are provided on our git repository.

\subsection{Baseline}
Initially our USV was controlled using a PID controller and a linear model of the system. This controller was not only hand-engineered for the environment but failed to reach velocities higher than 0.4 m/s. Hence, we compare the RL agent to a strong Model Predictive Path Integral (MPPI) controller~\cite{williams2016aggressive}, using a cost function similar to the reward of our agent. The MPPI is a monte-carlo-based controller that requires a model of the system.  Here this model is learned by a Neural Network using the real USV, but unlike our method which is relying solely on a 2D laser scanner, the MPPI uses the on-board RTK GPS and IMU to infer its state: linear velocity, and angular velocity. The Neural-Network used to model the system was pre-trained in simulation on 1.000.000 samples and finetuned on the real-system with 100.000 samples. Both trainings were done using the PER scheme and a grid-search was used to find the optimal parameters. We collected samples on the real USV as the MPPI would not converge on the real system when using the model acquired in simulation.

\begin{table*}
    \centering
    \vspace{4mm}
    \caption{Results (simulated and real system).
    Target velocity 1m/s, target distance from the shore 10meters.}
    \begin{tabular}{l||cc||cc|cc|c}
         & \multicolumn{2}{c||}{Simulated system}&\multicolumn{5}{c}{Real system} \\
         Representation & Ideal P2P & Ideal L2P  & Ideal P2P & Ideal L2P & Extra Drag P2P & Wind + Rain P2P & MPPI\\\hline
         Collisions & 0.0 & \textbf{0.0} & \textbf{0.0} & 0.5 & 0.0 & 0.0 & 0.0 \\
         Interventions & 0.0 & \textbf{0.0} & \textbf{0.0} & 1.2 & 0.0 & 0.0 & 0.7\\
         Velocity (m/s) & 1.1 $\pm$ 0.22 & \textbf{1.04 $\pm$ 0.18}  & \textbf{0.99 $\pm$ 0.17} & 0.98 $\pm$ 0.18 & 0.73 $\pm$ 0.12 & 0.97 $\pm$ 0.18 & 0.83 $\pm$ 0.19 \\
         Distance (m) & 9.8 $\pm$ 0.77 &  \textbf{9.9 $\pm$ 0.23} &\textbf{9.9 $\pm$ 1.67} & 11.1 $\pm$ 2.9& 10.32 $\pm$ 1.51 & 9.9 $\pm$ 1.77 & 9.5 $\pm$ 2.5\\
    \end{tabular}
    \label{tab:data_rep}
\end{table*}
\subsection{Evaluation}
To evaluate the performances of the different agents and the different representations, we record the following metrics:
    \textbf{1)} the number of collisions the agent has with the environment per 10 minutes;
      \textbf{2)} the number of times the agent fails to perform its shore-following task per 10 minutes; 
    \textbf{3)} the USV average linear speed $\pm$ its standard deviation;
    \textbf{4)} the USV average distance from the shore $\pm$ its standard deviation.
To further assess the robustness of the system we also study how the agent reacts when facing a hard situation, such as the hairpin bend, or when the agent starts too close or far from the shore. Furthermore, we deploy the agent in challenging weather conditions and also assess its behavior when its dynamic change.

\section{Result}

\subsection{Data Representation}
In this section we evaluate how the different data representations compare, both in simulation, and in real life. In these experiments we trained 2 types of agents: \textbf{``Projection To Projection'': P2P} (reconstructs laser projections from observed laser projections) and \textbf{``Laser To Projection'': L2P} (reconstructs laser projections from observed laser measurements). 
An \textbf{L2L} agent was also experimented, but not included for the sake of page limits. Its results are less good than the other agents', this could be tied to the difficulties of generating the values of a laser-scan in unstructured environment where most of the points are not reflected. 

As can be seen in Table~\ref{tab:data_rep}, the \textit{L2P} is the best performing approach in simulation. The laser-scans offer a high resolution input and provide farther information than the projection of the laser-scans it reconstructs. This allows the world model of the agent to be more accurate both instantaneously and over multiple time steps. However, the convergence of this model is more delicate than the \textit{P2P} one as the decoder of the \textit{L2P} agent must learn a more complex task. 
To alleviate this issue we found ideal to first train the \textit{P2P} agent and then use it to collect around 100 episodes for the \textit{L2P} agent. This allows to optimize the world model of the \textit{L2P} agent before letting it explore its environment. It has two benefits: it prevents an explosion of the value loss, and it reduces the total time required to train the agent.


On the real lake, the results are different: \textit{P2P} performs best. Overall, the \textit{P2P} agents have always performed their task perfectly: with over $8.2$ kilometers traveled during our experiments with this agent, the USV never required human assistance to fulfill its task. Regarding how well it kept its distance from the shore, with a mean of $9.9$ meters and a standard deviation of $1.67$ meters, the results are not drastically different from those in simulation. 
Concerning \textit{L2P}, the agent performs globally well but its performances are far from what it reached in simulation. This is due to the significant difference between real and simulated laser-scans. Nonetheless, both the \textit{P2P} and the \textit{L2P} agents are capable of maintaining a velocity of $0.99m/s$ even if they do not directly measure it.
This demonstrates that our agents, despite that they only have access to the laser-scans, are capable of estimating both the velocity of the system and their distance from the shore.

Finally, we observe that compared to other previously tried approaches for this task: a PID controller with a linear model, and the MPPI; the RL is the only one that is capable of following the shore at high speed. Furthermore, the RL is capable of following the shore in a single direction even-though its reward does not explicitly enforce it.


\subsection{Robustness}
\begin{figure}[h]
    \setlength{\tabcolsep}{2pt}
    \begin{tabular}{cc}
       \includegraphics[width=.45\linewidth]{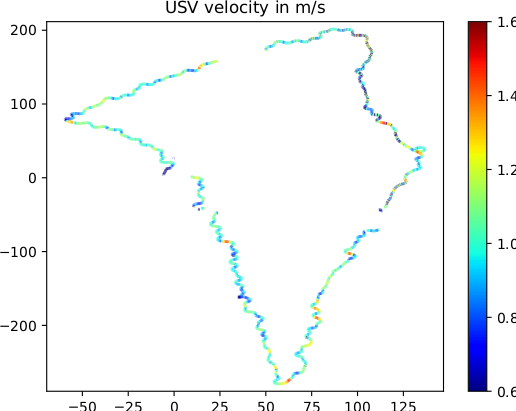}  &  \includegraphics[width=.45\linewidth]{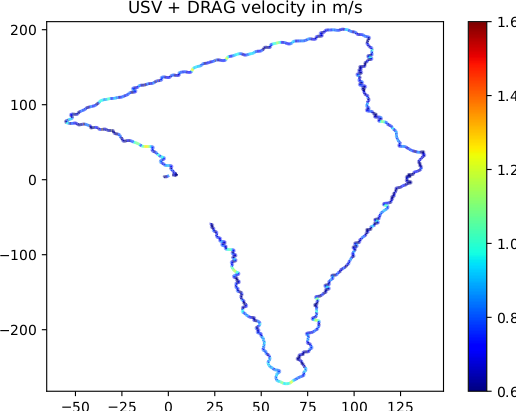}
    \end{tabular}
    \caption{The velocity of the USV around our lake. Left the USV without the swimming board, right the USV with the swimming board.}
    \label{fig:drag}
\end{figure}

To assess the robustness of our agent we attach to our USV a swimming board with a fin perpendicular to the forward direction. This device greatly increases the drag of the robot. We want to see how this contraption affects the behavior of the agent and if it is still capable of fulfilling its task. Because the \textit{P2P} agent was the most reliable in the previous field tests we only show its results here, on $2.2$ kilometers traveled during our experiments. As can be seen in Figure~\ref{fig:drag} the RL agent is capable of fulfilling its task despite the radical change in the system dynamics. However, its velocity is seriously impacted as the average speed of the USV with the extra drag is much lower. 

If we look at the commands sent to the system we can see that without the extra drag the agent uses 90\% of the available power to steer the boat whereas with the drag it uses 95\% of the power. This shows that it is trying to compensate for its lower speed but is probably limited by the power of the engines. While the agent does not reach the target velocity, it can be seen from Table \ref{tab:data_rep} that its behavior remains very close to the system without the swimming board. Most notably the agent maintained a similar distance to the shore, and never required assistance. In addition to the swimming board test, we also tested our approach under a strong wind and moderate rain. As can be seen in Table \ref{tab:data_rep}, the metrics of the model are not that different from the one acquired under ideal weather conditions. 
Furthermore, all the RL agents were capable to swiftly avoid the swans, and kept following the shore.

\subsection{Domain Randomization}
When we first trained the agents with DR their performances decreased. Initially, we had a very aggressive random spawn policy that would make the USV spawn close from the shore. The idea behind it is that if the boat spawns close from the shore then the agent learns how to quickly reach the correct distance and will know how to deal with complex situations. The main problem with this method is that with our settings, the episodes are not long enough, and the agents are not spending enough time navigating at the optimal distance from the shore which decreases their performances. 
To make matter worse agents trained without DR are as good as agents trained with DR to get out of hard situations. Which led us to reconsider the initial aggressive spawning strategy. Once this done, DR did provide some improvements on the L2P agent. Without DR the agent could lose track of the shore after large bushes, requiring human intervention to get back close to the shore. With DR we did not notice this behavior any longer which  indicates that DR improved the robustness of the system.
All in all, DR is delicate to tune, but with software like gazebo some settings can be changed very easily through ROS services to mimic how the behavior of the system can change: for instance changing the water density is an easy way to simulate battery voltage drop or weight changes. 


\subsection{Baseline Comparison}
In Table \ref{tab:data_rep} we also compare the different agents to MPPI, a state-aware model predictive controller. Across all metrics the \textit{P2P} agent performed better than the MPPI. This is interesting because the MPPI is not only more computationally expensive, but was also trained on the real system. Before being able to perform its task, we train a Multi-layer-perceptron to learn the dynamics of the system using its entire state (linear and angular velocities). While it is surely possible to tune the MPPI to improve its results, these same results show that our model can be trained in simulation, dropped on the lake and perform better than an algorithm that was tuned, trained, and optimized on the real task. Furthermore, our agent is capable of recovering from cases where its heading is facing the opposite direction we want it to navigate, whereas the MPPI does not. It means that during a survey the MPPI can, from time to time, change its forward direction, following the shore on its right side although it should follow it on its left side, or the other way around. Our agent does not. 

\section{Conclusion}

In this paper we teach a robot to solve a navigation task only based on a 2D laser scanner in an unstructured and natural environment by training it only in simulation. We demonstrate that the behavior learned by the robot is robust to changes in the dynamics, but also in the environment. Our agents performed well when facing moving obstacles, swans, or when we changed drastically their dynamics by increasing their drag. Moreover, despite that they are solely relying on a laser scanner the agents are capable of estimating their speed and their distance from the shore, making this solution applicable in many other environments. Future work will focus on transferring this RL agent from our USV to other robots such as Unmanned Ground Vehicles. Also, recent developments in model-free RL such as RAD~\cite{laskin2020reinforcement}, or DrQ~\cite{kostrikov2020image} will be investigated. Those approaches use image-augmentation as a regularization mechanism for SAC, enabling efficient learning in a low-data regime.


\bibliographystyle{IEEEtran}
\bibliography{biblio}

\end{document}